\documentclass[letterpaper, 10 pt, conference]{ieeeconf}  

\IEEEoverridecommandlockouts                              

\usepackage{textcomp, gensymb}
\usepackage{graphicx} 
\usepackage{amssymb}  
\usepackage{amsmath} 
\usepackage{microtype}

\usepackage{booktabs}
\usepackage{algorithm}
\usepackage{algorithmic}
\usepackage{colortbl}
\usepackage[x11names]{xcolor}
\definecolor{Gray}{gray}{0.85}
\definecolor{LightCyan}{rgb}{0.88,1,1}
\usepackage{comment}

\title{ \LARGE \bf SANDRO: a Robust Solver with a Splitting Strategy for  \\ Point Cloud Registration}

\author{ Michael Adlerstein, Jo\~ao Carlos Virgolino Soares$^{*}$, Angelo Bratta, Claudio Semini
\thanks{All authors are with the Dynamic Legged Systems (DLS) lab, Istituto Italiano di Tecnologia (IIT), Genova, Italy.}
\thanks{
Email: {\tt\small {name.surname@iit.it}}}
\thanks{$^{*}$ Corresponding author, \tt\small{joao.virgolino@itt.it}} 
\thanks{This work was supported by the European Union – NextGenerationEU, the
PNRR MUR Project PE000013 “Future Artificial Intelligence Research
(FAIR)”, and the PNRR MUR Project ECS00000035 “Robotics and AI for Socio-economic Empowerment (RAISE)”}
}

\begin{document}

\maketitle

\begin{abstract}
Point cloud registration is a critical problem in computer vision and robotics, especially in the field of navigation. Current methods often fail when faced with high outlier rates or take a long time to converge to a suitable solution.
In this work, we introduce a novel algorithm for point cloud registration called SANDRO\footnote{https://github.com/iit-DLSLab/SANDRO} (Splitting strategy for point cloud Alignment using Non-convex anD Robust Optimization), which combines an Iteratively Reweighted Least Squares (IRLS) framework with a robust loss function with graduated non-convexity. This approach is further enhanced by a splitting strategy designed to handle high outlier rates and skewed distributions of outliers. SANDRO is capable of addressing important limitations of existing methods, as in challenging scenarios where the presence of high outlier rates and point cloud symmetries significantly hinder convergence. 
SANDRO achieves superior performance in terms of success rate when compared to the state-of-the-art methods, 
demonstrating a 20\% improvement from the current state of the art when tested on the Redwood real dataset and 60\% improvement when tested on synthetic data.

\end{abstract}

\section{Introduction}
\label{sec:intro}

Point cloud registration is a crucial problem in computer vision, with applications in robot navigation, object reconstruction, and manipulation. In robot navigation, point cloud registration plays a key role in Simultaneous Localization and Mapping (SLAM), enabling robots to create and continuously update maps of their environment used in obstacle detection and avoidance, ensuring safe navigation. In object reconstruction, registration facilitates the merging of multiple partial scans to create accurate and complete 3D models, which is essential for various tasks such as 3D modeling, and industrial inspection~\cite{ye2020deep}~\cite{dastoorian2018automated}.

Despite significant advances, point cloud registration faces several difficulties in real-world scenarios. These include the presence of noise, outliers, partial overlaps between scans, and the computational complexity of processing large datasets. Outliers, in particular, pose a significant challenge, as incorrect matches between points can introduce large errors and significantly degrade the performance of traditional registration algorithms. The impact of outliers on accuracy makes robust registration methods an important area of research \cite{huang2022robust}.
In the likely case that a large part of the correspondences are incorrect, point cloud registration presents a ``chicken-and-egg" problem: the optimal transformation matrix can be determined if the true correspondences are known; conversely, accurate correspondences can be identified if the optimal transformation matrix is provided. However, solving these two aspects simultaneously is not straightforward \cite{huang2021comprehensive}.

Several techniques have been explored and are widely used for solving point cloud registration, ranging from traditional methods, such as the Iterative Closest Point (ICP) algorithm, to more advanced feature-based approaches. ICP is a well-established method that iteratively refines the alignment by minimizing the distances between corresponding points from two point clouds \cite{icp}. However, it requires a good initial guess to converge to a correct solution. In contrast, feature-based methods, such as RANSAC, extract and match features from point clouds to determine correspondences \cite{wang2017survey}, \cite{schnabel2007efficient}.

Feature-based methods often include a pre-processing step to filter out outliers before solving for the transformation matrix using the remaining correspondences. These pre-processing techniques range from simple approaches, such as the Tuple test, to more advanced methods, such as those in \cite{bustos2017guaranteed,zhou2018open3d}. However, even with effective filtering, some outliers may still persist, potentially causing the registration to fail. Therefore, a robust solver is crucial to ensure accurate registration when pre-processing does not fully eliminate all outliers.

\begin{figure}[t!]
    \centering
    \includegraphics[width=0.47\textwidth]{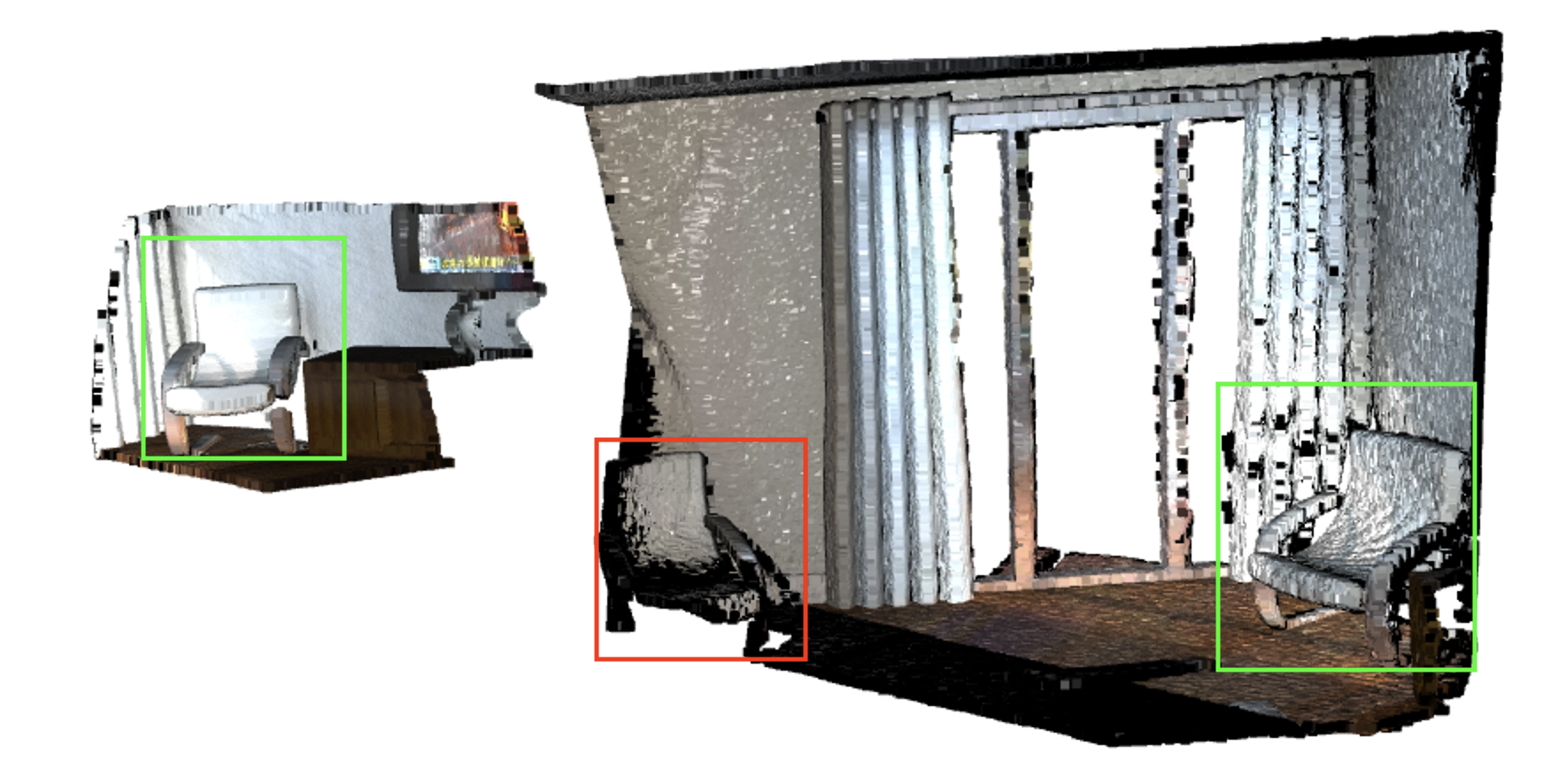}
    \vspace{-3mm}
    \caption{Example of a symmetric relationship with initial bias. Point cloud 1 (left), is contained within point cloud 2 (right). The two chairs in point cloud 2 cause current registration methods to fail due to the initial outlier distribution. The matching chairs are shown with green boxes.}
    \label{fig:bias}
    \vspace{-5mm}
\end{figure}

 Deep-learning-based methods leverage neural networks and have been used to extract more salient descriptors~ \cite{elbaz20173d,lu2019deepvcp}. However, they lack explainability and need to be pre-trained. In addition, they are often slow in inference, making them less suitable for online applications.

This paper proposes SANDRO (Splitting strategy for point cloud Alignment using
Non-convex anD Robust Optimization), a new solver for feature-based point cloud registration that delivers robust performance, even in scenarios with up to 95\% outliers. Our method can easily be integrated with widely used descriptors, such as Fast Point Feature Histograms (FPFH) features \cite{rusu2009fast} and does not require any initialization. 
Furthermore, SANDRO presents a novel splitting strategy that allows it to surpass current state-of-the-art solvers in scenarios where symmetries or uneven outlier distributions are present.

\begin{figure*}[!t]
    \centering
    \begin{minipage}[t]{0.5\textwidth}
        \centering
        \includegraphics[width=1\textwidth]{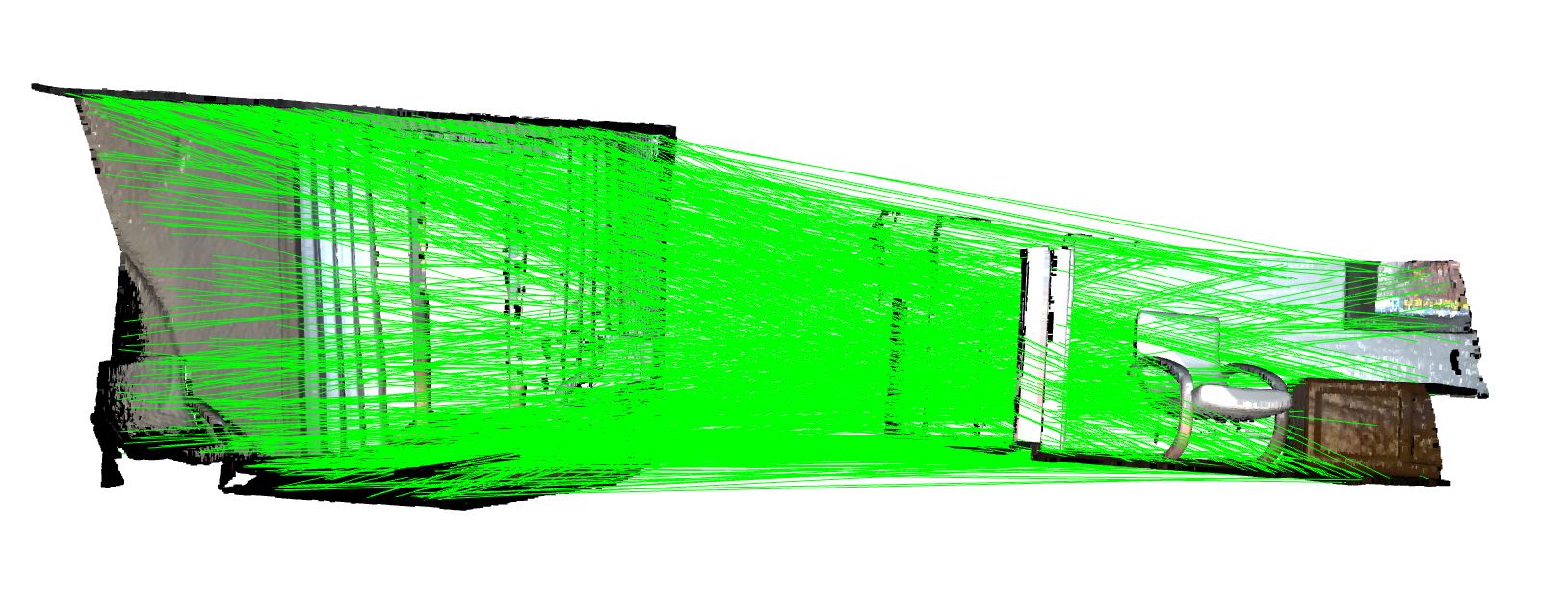}
        \label{fig:bias1}
    \end{minipage}%
    \hfill
    \begin{minipage}[t]{0.5\textwidth}
        \centering
        \includegraphics[width=0.8\textwidth]{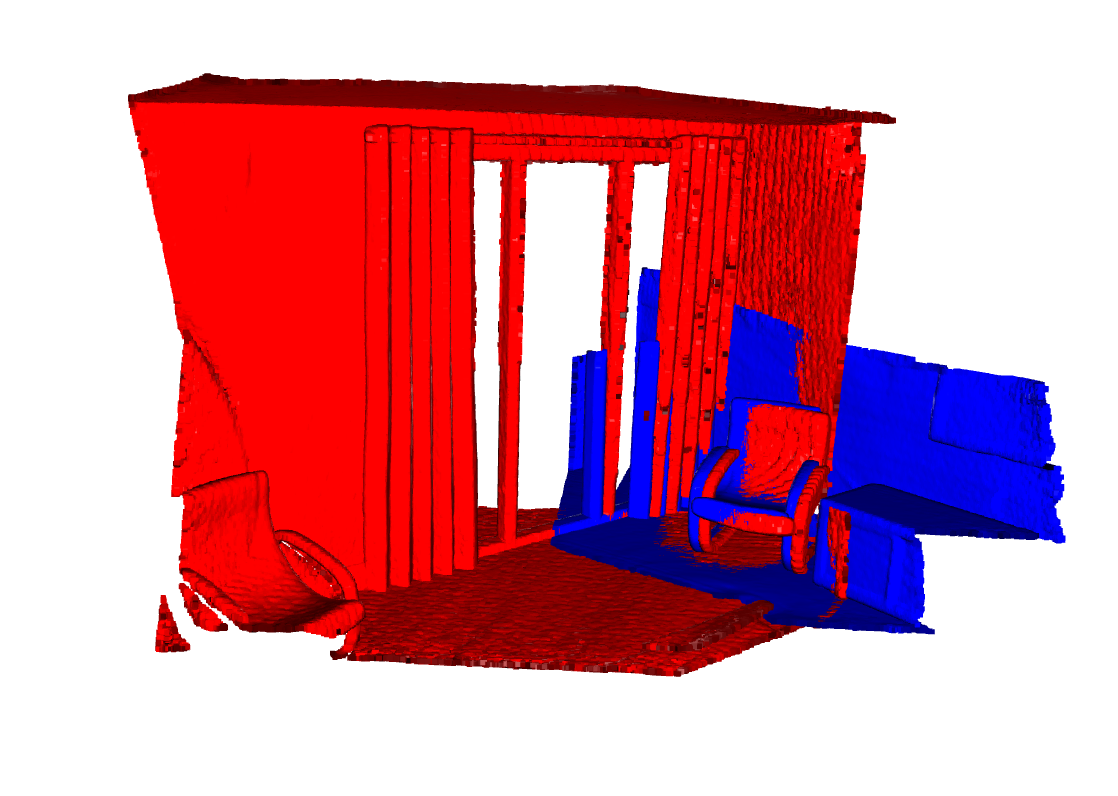}
        \label{fig:bias2}
    \end{minipage}
    \vspace{-10mm}
    \caption{Comparison of FPFH mutual matches between point clouds (left) and registered point clouds using SANDRO with the splitting strategy (right). Red and blue parts correspond to two different point clouds to be registered.
    In this example, the presence of two similar chairs in the point clouds causes the outlier distribution to be highly skewed, preventing convergence with traditional methods. The splitting strategy aims to break the initial bias in distribution by performing registration on independent subsets. The combined point cloud on the right shows the final registration of SANDRO with 4 splits.
    }
    \label{fig:combined}
    \vspace{-5mm}
\end{figure*}
\subsection{Contribution}
Although existing methods perform well on various datasets, many solvers struggle when dealing with symmetric distributions among correspondences. For instance, in a scenario with two chairs, as shown in Fig. \ref{fig:bias}, previous methods fail to converge to the optimal solution due to the bias introduced by the non-random distribution of outlier correspondences.

This paper introduces a novel solver that addresses this issue by applying Iteratively Reweighted Least Squares (IRLS)~\cite{Chartrand2008IterativelyRA} combined with the Geman-McClure~\cite{geman} robust function. A key innovation in our approach is the splitting strategy, where we operate on smaller independent subsets of the point cloud. By focusing on smaller portions, our method effectively reduces the initial outlier bias and significantly improves convergence toward the optimal solution.

Additionally, the proposed method runs in the order of hundreds of milliseconds when using the splitting strategy, and tens of milliseconds with no splits, allowing its implementation in scenarios that require online operation, such as SLAM. The algorithm, which combines Graduated Non-convexity (GNC) with a Geman-McClure robust loss and splitting, demonstrates state-of-the-art performance on both real and synthetic datasets, even with outlier rates of up to 95\%.

In summary, the main contributions of this work are the following:

\begin{itemize}

\item A novel point cloud registration solver that incorporates GNC with a Geman-McClure robust loss function is used to change the convexity of the optimization problem in order to iteratively filter out more outliers and converge to the non-convex global minimum. Our approach demonstrates improvements of up to 60\% in terms of success rate  when tested on a synthetic dataset with high outlier rates.
  
  \item A splitting strategy which demonstrates a 20\% improvement in the registration's success rate compared to previous methods in real datasets where symmetries are present.
  
\end{itemize}

These two features make SANDRO optimal for operating in structured scenarios where symmetries and initial distribution biases are present, such as shown in Figs.~\ref{fig:bias} and~\ref{fig:combined}.

The rest of the paper is organized as follows. Section II gives a literature review, Section III presents the proposed method, Section IV shows the results, and Section V draws conclusion and future work.

\section{Literature review}
Feature-based point cloud registration detects 3D keypoints and matches them using feature descriptors. Hand-crafted descriptors such as FPFH are fast but often produce high outlier rates, sometimes over 90\%. In contrast, neural network-based methods, such as the one proposed by Li et al.~\cite{li2020end}, generate more accurate descriptors, but at the cost of longer inference times, limiting their efficiency in real-time applications.

After matching the keypoints, methods as those proposed by Horn~\cite{horn1987closed} and Arun~\cite{arun1987least} provide an optimal maximum likelihood solution for computing the transformation between point clouds. These methods work well under the assumptions that the data is free from outliers, and the noise  between matches follows an isotropic zero-mean Gaussian noise.

To account for the high outlier rate, several methods have been explored, with the most widely used being RANSAC and its variations~\cite{bolles1981ransac}. RANSAC aims to align subsets of corresponding points until a loss threshold is met~\cite{yang2022ransacs}. The goal is to randomly select a subset of points that does not include outliers. This approach is popular due to its simple implementation and availability in open-source libraries, e.g., Open3D \cite{zhou2018open3d,choi1997performance}. While RANSAC can theoretically achieve optimal results given enough time, it tends to fail or become very slow when dealing with a large number of outliers. 

The Branch and Bound (BnB) method has been successfully applied to solve optimization problems by iteratively dividing the search space into smaller subspaces, calculating error bounds, and pruning those subspaces that cannot contain the optimal solution. In the context of point cloud registration, Bazin et. al. \cite{bazin2013globally} used a BnB-based method to find the rotation error, and improvements were proposed by Aoki. et. al.~\cite{aoki20243d} to improve the computational speed.
Although BnB guarantees a numerically optimal solution, it does so at the cost of requiring polynomial time, making it computationally expensive for larger problems and scenarios where local minima are present. 

Hitchcox and Forbes \cite{hitchcox2022mind} observed that most loss functions assume that the residuals follow a Gaussian-like distribution with a mode of zero. However, in many nonlinear least-squares problems, the residuals are actually defined as the norm of a multivariate error, which produces a Chi-like distribution with a nonzero mode. This 
demonstrates that relying too strongly on an isotropic zero mean outlier distribution can lead to poorer results. 

Non-minimal solvers, often combined with robust estimators, are designed to leverage data redundancy, making them less sensitive to measurement noise compared to minimal solvers \cite{khoo2016non}. M-estimators modify the nonlinear least-squares cost function by incorporating robust loss functions, reducing the influence of outliers during optimization \cite{santana2024proprioceptive}.

M-estimators can also be used with GNC, which optimizes the robust costs by transitioning from convex to non-convex surrogate functions. This approach helps to avoid local minima during optimization.
Jung et al. \cite{jung2024adaptive} explored the performance of various robust loss functions, both with and without GNC, and demonstrated that using GNC significantly improves the robustness of point cloud registration.

M-estimation is commonly implemented in an IRLS framework. IRLS and its variants have demonstrated strong performance in point cloud registration and have been proven to offer robust convergence properties, as shown by Peng et al. \cite{peng2023convergence}.
A well-known example of M-estimation with GNC is Fast Global Registration (FAST) by Zhou et al. \cite{zhou2016fast}. FAST combines an optimization approach based on the Jacobian matrix with a Geman-McClure robust loss function, allowing it to iteratively refine the solution and minimize alignment errors between point clouds.
In FAST, the GNC helps find an initial alignment in convex regions, avoiding local minima before moving into non-convex areas. The use of the Jacobian matrix speeds up gradient computation and improves convergence. Due to its speed and robustness, FAST is widely used in real-world applications and is implemented in libraries, such as Open3D \cite{zhou2018open3d}, making it well-suited for large-scale registration tasks.

Yang et. al proposed ADAPT~\cite{yang2020graduated}, a GNC-based optimizer which showcase state-of-the-art performance when tested on a variety of datasets. ADAPT was combined with rotation and translation invariant feature extractions~\cite{yang2020teaser}.
ADAPT works by means of Semi-definite programming and it has been shown to be resilient to up to 90\% outliers in a variety of computer vision applications.
In a practical implementation, ADAPT uses Truncated Least squares alongside GNC and Black-Rangarajan duality to solve the registration problem. 
The optimizer has been shown to reach certifiably optimal solutions, making it the current state-of-the-art solver for point cloud registration.  

Despite promising advancements in point cloud registration,  current solvers still display poor performance when prefiltering is not applied. Furthermore, even when prefiltering is present, inlier points can get classified as outliers due to the initial distribution of the data. Our method aims to overcome this problem by including all the points in the optimization process, including all outliers, and by implementing a splitting strategy to ease the optimization.

\section{Method}
\label{sec:method}
\begin{figure}[h]
    \centering
    \includegraphics[width=0.45\textwidth]{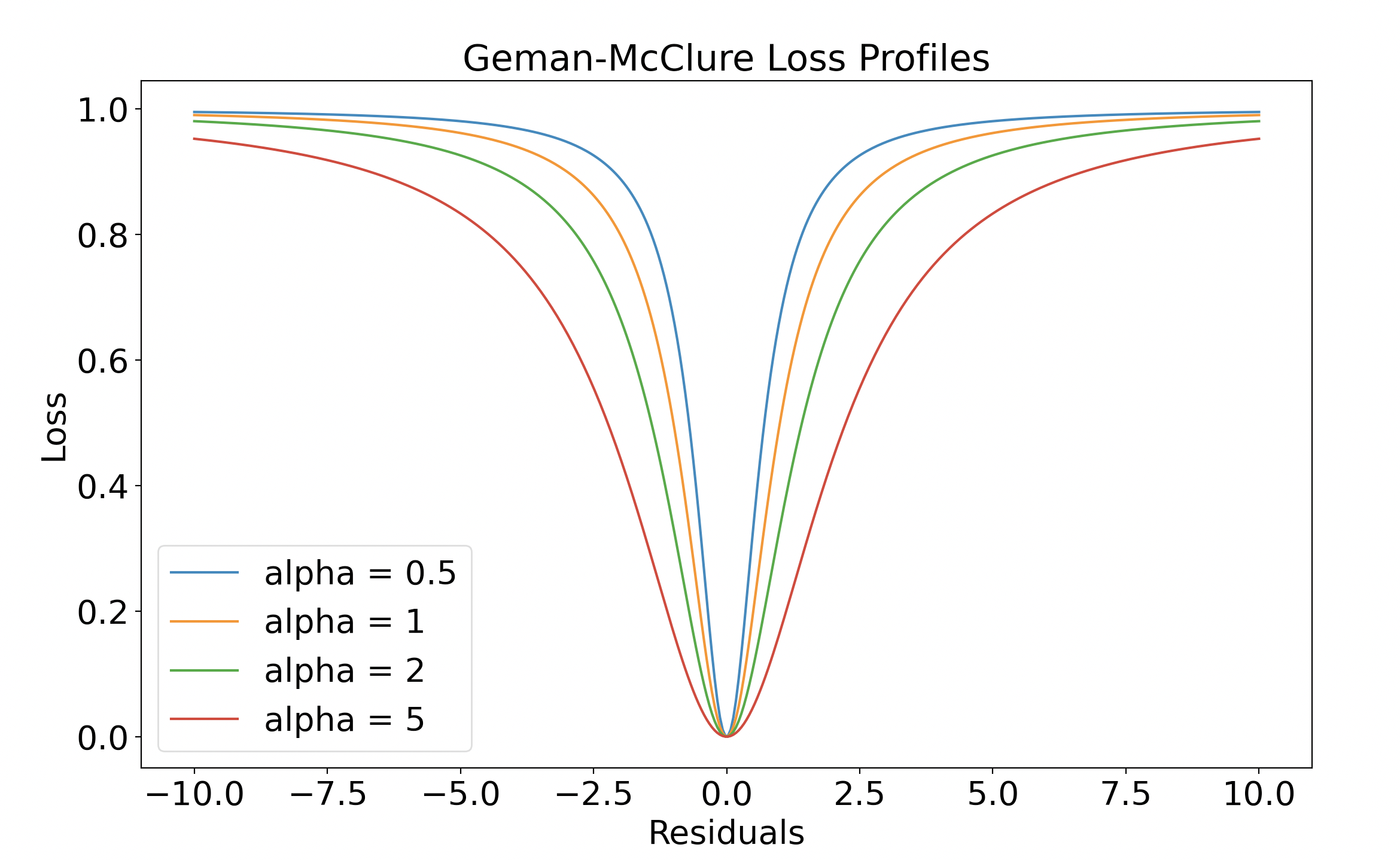}
    \vspace{-2mm}
    \caption{GNC applied to the Geman-McClure loss for different values of the decay parameter $\alpha$.}
    \label{fig:gnc_gmk}
\vspace{-3mm}
\end{figure}

The proposed method incorporates the Geman-McClure loss, alongside a GNC with exponential decay (see Fig.~\ref{fig:gnc_gmk}) in an IRLS framework. Additionally, the optimizer utilizes a splitting strategy, which allows it to explore several solutions obtained by taking subsets of the point cloud. 
This approach aims to break the initial skewed outlier distribution into smaller independent subsets which resemble more closely a zero mean distribution. Our splitting strategy offers an advantage over methods that take the entire outlier distribution. Traditional methods fail in the cases where symmetries are present, such as in Fig.~\ref{fig:bias}. 
SANDRO accounts for initial geometric relations within the point cloud that might otherwise cause premature convergence to local minima during the GNC evaluation. By doing so, the optimizer effectively reduces the risk of getting trapped in sub-optimal solutions. 

\vspace{-3mm}

\subsection{SANDRO's algorithm}
\label{sec_algorithm}
Given two point clouds, $\mathbf{P} = \{\mathbf{p}_i\}$ and $\mathbf{Q} = \{\mathbf{q}_i\}$ with $N$ matching points, we formulate the registration problem as a robust least square problem: 
\begin{equation}
\displaystyle{\min_{\mathbf{T}}}\quad
\sum_{i}^N \rho(\mathbf{r}_i(\mathbf{T}))
\label{eq:registration}
\end{equation}

\noindent where $\mathbf{T} \in SE(3) $ is the homogeneous transformation matrix between $\mathbf{P}$ and $\mathbf{Q}$, and $\mathbf{r}_i \in \mathbb{R}^3$ is the residual $\mathbf{r}_i~= \mathbf{p}_i - \mathbf{T} \mathbf{q}_i $. For brevity of notation, the dependence of the residuals on $\mathbf{T}$ is omitted.
The function $\rho$ $ : \mathbb{R}^3 \to \mathbb{R}$ is a robust cost function. As already mentioned, in our approach we combine the Geman-McClure function with the GNC approach, so \eqref{eq:registration} is rewritten as:

\begin{equation}
\displaystyle{\min_{\mathbf{T}}}\quad
\gamma =
 \sum_{i}^N E_i(\mathbf{T}) =  \sum_{i}^N
\frac{\alpha \| \mathbf{r}_i \| ^2}{\alpha + \| \mathbf{r}_i \| ^2}
\label{eq:gnc}
\end{equation}

The function $E_i(\mathbf{T})$
 in (\ref{eq:gnc}) corresponds to the GNC surrogate function of the Geman-McClure function~\cite{yang2020graduated}, and is solved by iteratively reducing the decay parameter $\alpha$. We used an exponential decay, as it allows the loss function to transition progressively from a convex to a less convex profile, enabling smoother convergence toward the global minimum. With each iteration of the IRLS, the optimization problem becomes locally convex around the current solution, making it easier for the algorithm to move closer to the global optimum while avoiding poor local minima.

Equation~\eqref{eq:gnc} can be efficiently solved by applying the IRLS framework and then implementing the weighted SVD algorithm in~\cite{sorkine2017least}. Thus, the minimization problem is rewritten as:

\vspace{-3mm}
\begin{equation}
\displaystyle{\min_{\mathbf{T}}}\quad
\sum_{i}^{N} \frac{1}{2} w_i \| \mathbf{r}_i \|^2
\label{eq:irls}
\end{equation}

The weights $w_i$ are updated with the gradient of the robust cost function  $E_i(\mathbf{T})$ with respect to $\| \mathbf{r} \| $

\begin{equation}
w_i = \frac{1}{\| \mathbf{r}_i \| }
\frac{\partial E(\mathbf{T})}{\partial \| \mathbf{r}_i \| } 
= \frac{1}{2}  \frac{1}{\| \mathbf{r}_i \| } 
\frac{2 \alpha^2 \| \mathbf{r} \|^2 } {(\alpha + \| \mathbf{r}_i \| ^2)^2 }
 =
\frac{\alpha^2 }{(\alpha + \| \mathbf{r}_i \|^2)^2}
\label{eq:weights}
\end{equation}

\noindent allowing for the adaptive reweighing during the IRLS process. As $\alpha$ decreases, the weighting function becomes sharper, resulting in points with smaller residuals to have a higher weight. It is important to note that SANDRO does not explicitly discard outlier points during the optimization. Instead, it updates them to have a lower impact on the final transformation matrix.

To summarize, in our algorithm we solve \eqref{eq:irls}, we reduce $\alpha$ with an exponential parameter and we update the residual $\mathbf{r}_i$ and consequently the weights~\eqref{eq:weights} using the transformation matrix $\mathbf{T}$ previously computed. These steps are performed until no reduction is obtained between two consecutive steps for the Geman-McClure function~\eqref{eq:gnc}.
Furthermore, our optimizer does not require a good initial starting point, which significantly enhances its usability in practical applications where obtaining such a starting point is challenging. 

\subsection{Splitting strategy}

As already mentioned, similar approaches that utilize the entire
set of matched points are often unable to converge due to the initial skew in outlier correlations. To improve performance, we introduced a splitting strategy, where the initial point cloud is divided into equal parts called sub-clouds. The number of splits can be adjusted as a hyper-parameter dependent on the size of the point cloud. This approach helps the algorithm to focus on smaller sections without being influenced by non zero-mean outlier distributions, which could otherwise hinder the IRLS algorithm’s ability to converge. 

The steps described in Sec. \ref{sec_algorithm} are performed independently on each sub-cloud. Once the optimization step for each sub-cloud is completed, the transformation with the lowest loss is selected as the best estimate to aligning the entire point cloud.

SANDRO with splitting strategy is summarized in Algorithm 1.

\begin{algorithm}
\begin{algorithmic}[1]
\REQUIRE Point clouds $\mathbf{P} = \{\mathbf{p}_i\}$ and $\mathbf{Q} = \{\mathbf{q}_i\}$, number of correspondences $N$ , initial transformation $\mathbf{T}_0$, parameter $\alpha$, tolerance $\epsilon$ , decay rate $\beta$
\STATE Initialize $\mathbf{T} \leftarrow \mathbf{T}_0$
\STATE Define the number of splits  $s$
\STATE split matches in  $\mathbf{P}$ and $\mathbf{Q}$ by $s$ equal sub clouds

\REPEAT 
    \STATE Compute residuals $\mathbf{r}_i \leftarrow  \mathbf{p}_i - \mathbf{T} \mathbf{q}_i $ for all $i$
    \STATE Compute weights $w_i \leftarrow \frac{\alpha^2}{(\alpha + \| \mathbf{r}_i \|^2)^2}$ for all $i$
    
    \STATE Compute robust loss $\gamma \leftarrow \sum_{i} ^{N/s} \frac{ \| \mathbf{r}_i \|^{2}}{(\alpha + \| \mathbf{r}_i \| ^2)}$

    \STATE Formulate and solve the weighted least squares problem:
    \[
    \mathbf{T} \leftarrow \arg\min_\mathbf{T} \sum_i^{N/s}  \frac{1}{2} w_i \| \mathbf{r}_i \| ^2
    \]  
    \STATE Update the transformation $\mathbf{T}$
    \STATE Update the decay parameter $\alpha \rightarrow \alpha \times\beta$
    \STATE Check for convergence:
    
    \[
    \Delta \gamma \leftarrow \| \gamma - \gamma_{\text{prev}} \|
    \]

\UNTIL $ \Delta \gamma  < \epsilon$ where $\epsilon$ the termination criterion 
\RETURN $\mathbf{T}$ for the sub cloud with the smallest robust loss.

\caption{SANDRO's pipeline}
    
\end{algorithmic}

\end{algorithm}

\vspace{-3mm}
\section{Results}

The proposed method was evaluated on both the Redwood dataset~\cite{choi2015robust} and on synthetic point clouds obtained in our lab.
The Redwood dataset contains a set of point clouds obtained from an indoor setting, which allow us to evaluate the performance of the different algorithms in highly structured scenarios. 

Our method was compared with state-of-the-art registration techniques, such as ADAPT and FAST. To generate a substantial number of matches between the two point clouds, FPFH features with mutual best matching scores were utilized. Although FPFH features are extremely fast to compute, they tend to produce a significant number of outlier correspondences, making them an excellent benchmark for assessing the effectiveness of the solvers.

\subsection{Evaluation on Redwood Dataset}
\label{sec:redwood}
\begin{figure}[htbp]
    \centering
        \centering
        \includegraphics[width=0.42\textwidth]{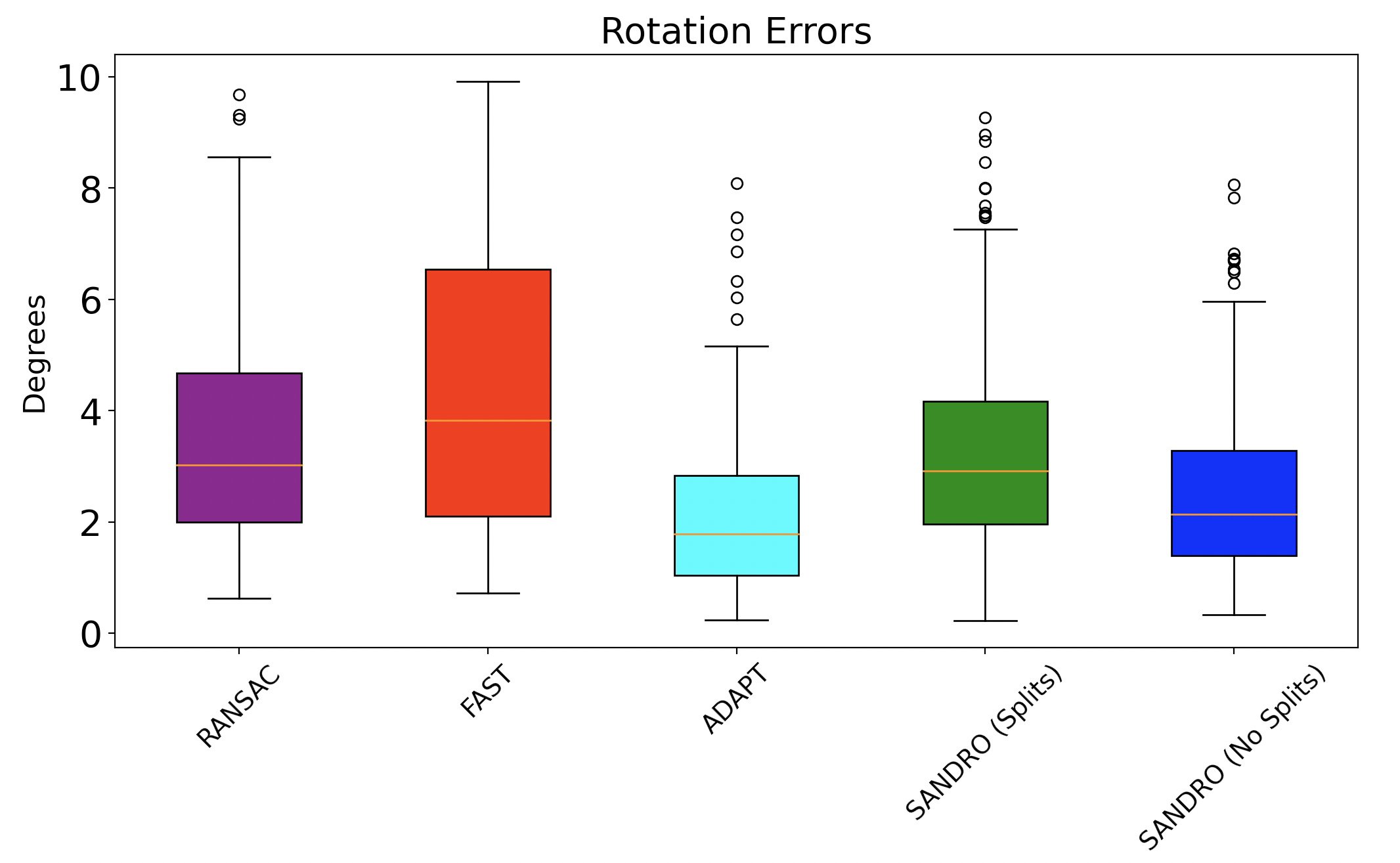}
        \includegraphics[width=0.435\textwidth]{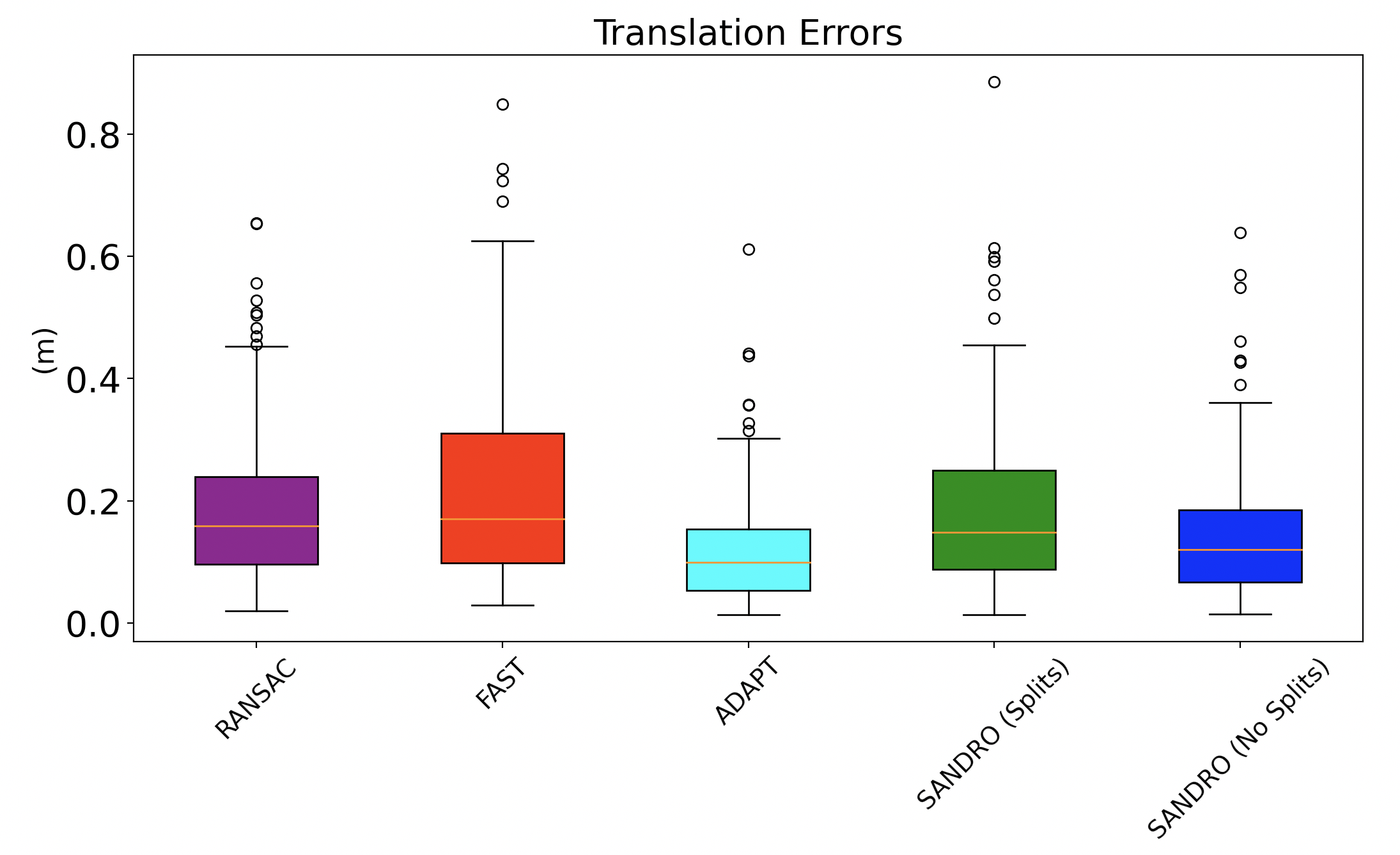}
        \vspace{-3mm}
    \caption{Comparison of Rotation and Translation Errors, displayed only for the point cloud pairs which achieved a successful registration in the Redwood dataset.
    }
    \label{fig:errors_comparison}
\end{figure}

For the Redwood dataset, the full set of matching point clouds was provided along with the correct transformation matrix. The dataset includes both simple scenarios, where the scans are very similar with minimal deviations, and more challenging ones, where point cloud pairs have low overlap and symmetric features, making registration difficult. To reduce redundant data, the point clouds were downsampled using a voxel size of 0.05. The points with the reciprocal best FPFH scores were then selected as representative matches for the registration process.
The results showcasing the performance of different methods are shown in Fig.~\ref{fig:errors_comparison}.

\begin{table}[h!]
\vspace{-3mm}
    \caption{Comparison of discarded matches and success rate between different methods. SANDRO with 4 splits successfully registers the most point clouds}
\begin{center}
\begin{tabular}{ |c|c|c| } 
 \toprule[0.4mm]
			\textbf{method} & {\textbf{discarded}} & \textbf{Success rate (\%)} \\ 
			\midrule
 RANSAC \cite{choi2015robust} & 214 & 35  \\ 
 FAST \cite{zhou2016fast} & 226 & 31 \\ 
 ADAPT \cite{yang2020graduated} & 187 & 43 \\ 
 SANDRO (4 splits) (ours) & 124 & 62 \\ 
SANDRO (no splits) (ours) & 181 & 45 \\ 

 \hline
\end{tabular}
\end{center}
\label{success_table}
\vspace{-3mm}
\end{table}

Successful registrations were defined as those with a rotation error $ \leq 10^\circ$ and a translation error $\leq 1\,\text{m}$.
These values were chosen by considering a SLAM application where the data is later fed into a pose graph optimization module. While pose graph optimization is a powerful technique that can correct odometry error, it is often unable to recover the true pose when the errors 
are too high, or when there are a large number of outliers.
Figure~\ref{fig:errors_comparison} shows that ADAPT and SANDRO (no splits) outperform RANSAC and FAST in terms of error values for both rotation and translation, considering successful registrations. 
Since RANSAC can technically converge to the best possible solution given an infinite time, 1000 iterations were selected as the maximum number in order to establish an upper bound. The average time for 1000 iterations of RANSAC was approximately 0.2 seconds, which is an order of magnitude slower than the other methods, including FAST, ADAPT, and SANDRO (no splits). 
We chose 4 splits for a good trade-off in terms of speed and success rate. The splitting strategy in SANDRO increases the computation time linearly; with 4 splits, the execution time becomes comparable to that of RANSAC with 1000 iterations. Furthermore, it was observed that 4 splits yielded the best results in term of success rate for alignment. 

The success rate of the different registration algorithms on the Redwood dataset is shown in Tab. \ref{success_table}. It is noticeable that incorporating the splitting strategy in SANDRO significantly improves the results compared to the other methods tested. This suggests that the splitting strategy is effective in handling non-uniformly distributed outliers.

This conclusion is further supported by comparing the results of SANDRO (no splits) to those of ADAPT. While both SANDRO and ADAPT minimize \eqref{eq:registration}
using GNC and IRLS, the Geman-McClure loss function used in SANDRO produces results similar to the Truncated Least Squares method used by ADAPT, with only a 2\% improvement in success rate. However, when the splitting strategy is applied, SANDRO shows a significant improvement over both ADAPT and SANDRO (no splits).

\vspace{-1.5mm}

\subsection{Evaluation on Synthetic Dataset}
To evaluate the registration algorithms in a controlled setting, we captured a point cloud of a standing person in our lab containing 88669 points, which was downsampled using a voxel size of 0.02. The lab-captured point cloud shows a loose symmetry between the left and right sides of the body, primarily due to the positioning of the subject.

To create a synthetic dataset from the captured point cloud, we generated outliers by projecting points from the target point cloud onto a unit sphere. This process was followed by a random rotation and translation in order to assess different starting scenarios between the source and target point clouds.

\begin{figure}[!b]
\vspace{-6mm}
    \centering
        \centering
        \includegraphics[width=0.35\textwidth]{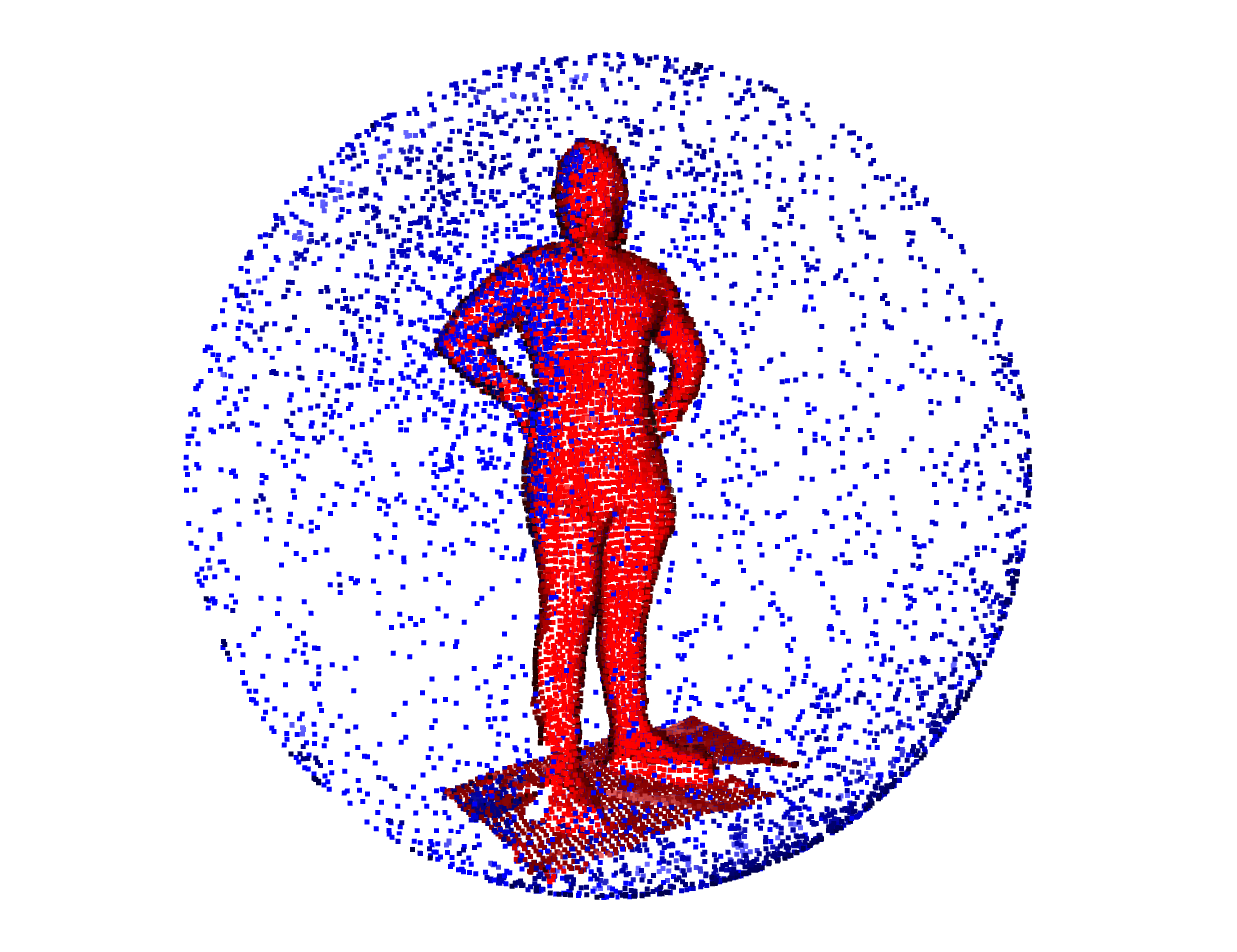}
        \vspace{-4mm}
        \caption{Aligned point clouds from the synthetic dataset with 50\% outlier rate, representing a person standing. The points in red correspond to the source cloud, and the points in blue are the target cloud with outliers.}
        \label{fig:noisy sandro}
\end{figure}

An example of the result of SANDRO is shown in Fig.~\ref{fig:noisy sandro}. It can be seen that SANDRO successfully recovers the correct transformation between the source and the target point clouds. The remaining inlier points were perfectly matched to maintain precise control over the percentage of outliers in the dataset. Each registration algorithm was tested 40 times with different translations to ensure that a statistically significant amount of data was collected for evaluation.

The synthetic dataset was evaluated by comparing the rotation error between the estimated and generated transformation matrices. Successful attempts were classified to be the ones with a rotation error $\leq 10 \degree $  and translation error $\leq 1$ m.
\begin{figure}[!t]
\centering
\includegraphics[width=0.43\textwidth]{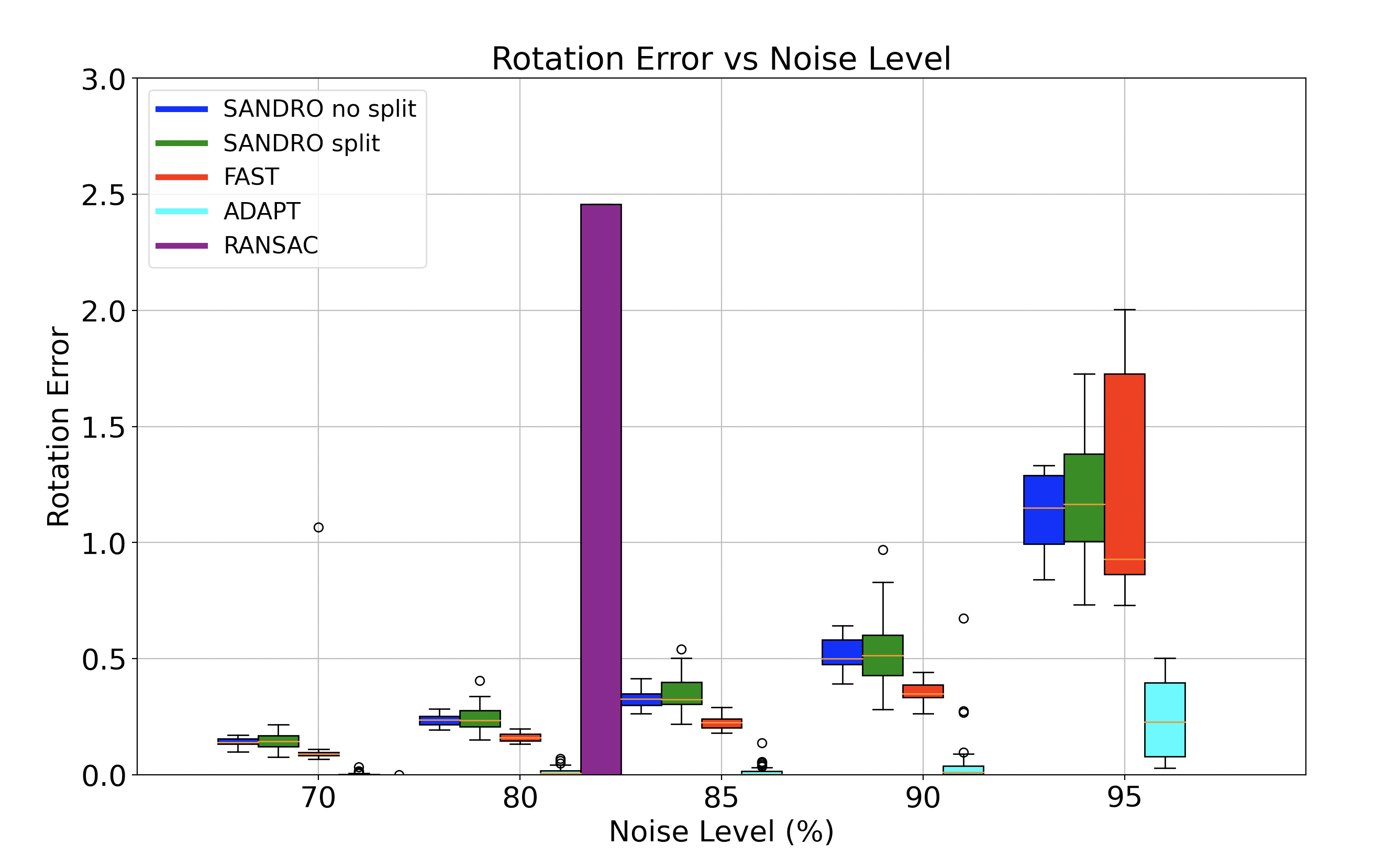}
\vspace{-3mm}
    \caption{Rotation errors in the presence of different outlier rates (or ``noise level") for five point cloud registration methods.}
    \label{fig:rotation_error_outlier}
    \vspace{-2.5mm}
\end{figure}
Figure~\ref{fig:rotation_error_outlier} shows that for outlier rates of up to 80\%, the rotation errors amongst the different methods remain very similar. After 80\%, all methods except ADAPT increase at similar rates and reach a rotation error of $\approx 1 \degree$ at 95\% outlier rate. 
\begin{figure}[!t]
\centering   
\includegraphics[width=0.44\textwidth]{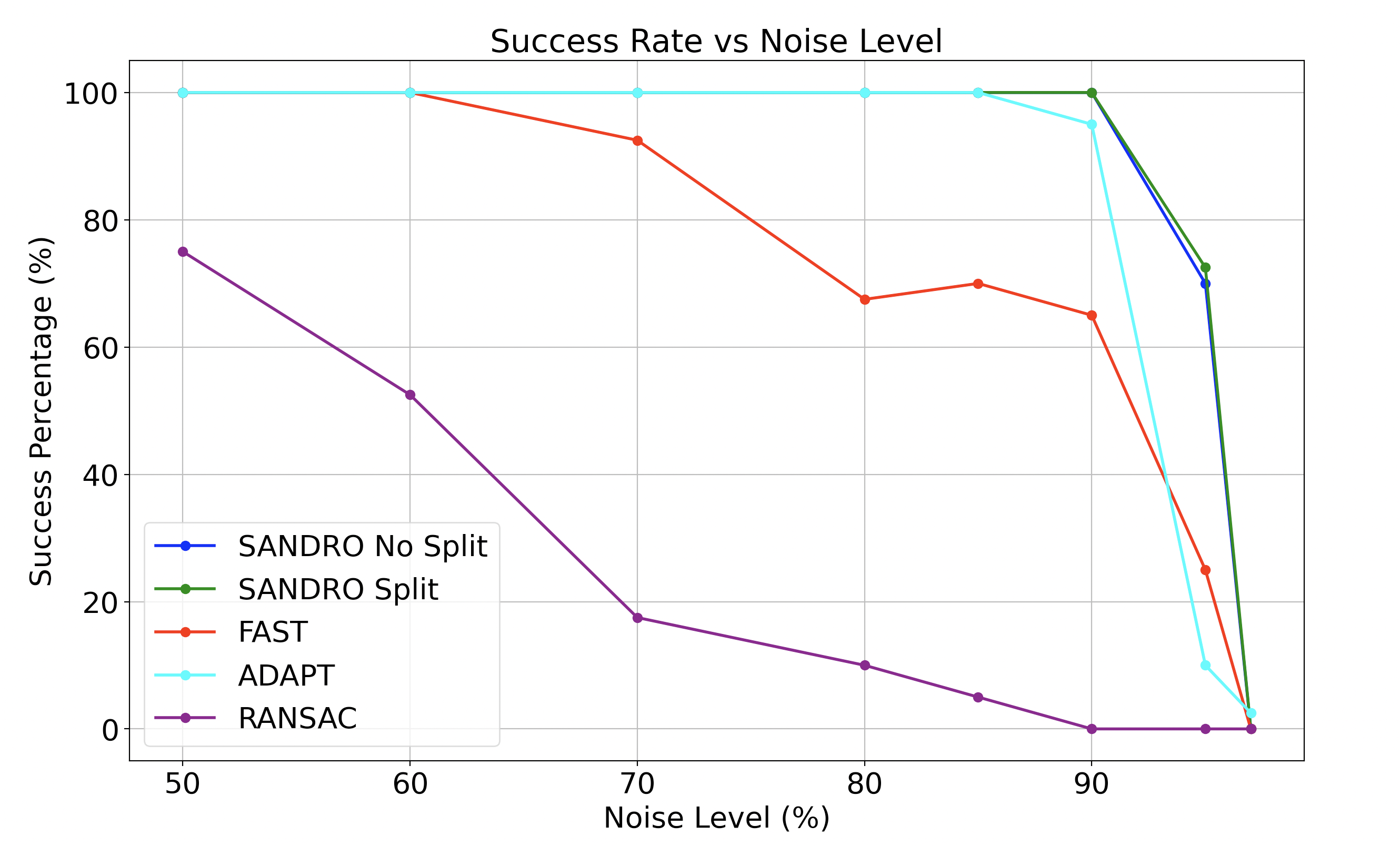}
\vspace{-3mm}
    \caption{Comparison of success rate with different outlier rates (or ``noise level") for five point cloud registration methods.}
    \label{fig:success_rate}
    \vspace{-5mm}
\end{figure}
Figure \ref{fig:success_rate} illustrates the success percentage of different methods at varying percentage of outlier rates. As shown, both versions of SANDRO, splitting and no splitting, demonstrate robust performance, successfully handling outlier rates of up to 95\%. In contrast, ADAPT and FAST show much lower tolerance at 95\% of outliers, with ADAPT maintaining a success rate of approximately 10\%, and FAST achieving around 25\%. This clear disparity highlights the superior robustness of the SANDRO methods in environments with high levels of noise.

Interestingly, there was no significant difference between the results of SANDRO with and without splitting. This lack of distinction could be due to the distribution of the synthetic dataset’s outliers. The uniform splitting of the point cloud likely resulted in sub-clouds with similar distributions, meaning the split sub-clouds did not provide any advantage over processing the entire cloud. As a result, both methods converged to similar performance metrics, suggesting that in this specific synthetic scenario, the splitting did not contribute meaningfully to the overall robustness.

As done in Sec. \ref{sec:redwood}, we selected a maximum number of 1000 iterations for RANSAC. As expected, it failed to reach a suitable solution when the outlier rates exceeded 60\%, due to the stochastic nature of the algorithm.

\subsection{Discussion}
The number of splits on the Redwood dataset was chosen to be the one showcasing the best results. We observed that after 4 splits the performance of the registration dropped for point clouds averaging around 1600 points. This can be attributed to the sub clouds not having a high enough number of inlier points, thus preventing it from converging.

In Fig. \ref{fig:rotation_error_outlier}, ADAPT is shown to have a lower rotation error throughout the different outlier rates compared to the other tested methods. This performance is most likely caused by the use of the Black-Rangarajan formulation used to perform the annealing of the line processes \cite{black1996unification}. This method proves to be more robust in terms of diminishing the rotation error with the trade-off of a lower success rate, as seen by the performance of ADAPT in the real and synthetic dataset.

Furthermore, the splitting strategy was observed to increase the time by a factor of ten. This increase in time was not caused directly by the optimization within the sub-clouds, instead it was caused by the non-optimized generation of the sub-cloud in the pre-processing of the matches. In the future, an optimized splitting function would be able to achieve a better computational time.
Overall, the method has demonstrated robust performance, consistently converging to high-quality solutions even in the presence of a high outlier rate. This makes it particularly suitable for complex computer vision tasks, such as 3D reconstruction, where traditional methods often struggle.

\section{Conclusion}

In this paper, we introduced a novel state-of-the-art solver capable of performing point cloud registration, even with high outlier rates on both real and synthetic datasets. Building on the GNC approach with the Geman-McClure cost function, we demonstrated that splitting the point cloud significantly improves registration success in scenarios with symmetries. This is shown in Fig. \ref{fig:combined}, where, on top of the low overlap between the point clouds, the majority of the matches does not follow the same distribution, resulting in a very skewed symmetrical outlier distribution.

When tested on the Redwood dataset, SANDRO, with 4 splits, outperformed previous state-of-the-art methods such as ADAPT, achieving better registration success rates (38\% discarded versus 57\% discarded by ADAPT). Additionally, in tests on the synthetic dataset, both variations of SANDRO noticeably outperformed other methods, achieving over 60\% success in registrations  with 95\% outlier rates, highlighting its robustness in scenarios with a high outlier rate.

As future works, the splitting strategy used in SANDRO could be further refined by determining the optimal number of splits required to account for different data distributions. One approach to achieve this would be to analyze the distribution of matched pairs and use that information to determine the optimal number of splits.
Additionally, a novel pre-processing methodology could be developed to better utilize the splitting strategy presented in this paper.

\bibliographystyle{IEEEtran}
\bibliography{main}

\end{document}